\documentclass{article} % ICLR 2025 style
\PassOptionsToPackage{table}{xcolor} % pre-empt option clash, xcolor is loaded by a dependency
\usepackage{iclr2025_conference,times}
\iclrfinalcopy

\usepackage{amsmath,amsfonts,bm}

\def\eqref#1{equation~\ref{#1}}
\def\1{\bm{1}}

\DeclareMathAlphabet{\mathsfit}{\encodingdefault}{\sfdefault}{m}{sl}
\SetMathAlphabet{\mathsfit}{bold}{\encodingdefault}{\sfdefault}{bx}{n}

\usepackage{placeins}
\usepackage{xcolor}
\usepackage{hyperref}
\hypersetup{hidelinks}
\usepackage{url}
\usepackage{booktabs}
\usepackage{multirow}
\usepackage{tabularx}
\usepackage{graphicx}
\usepackage{amsmath,amssymb}
\usepackage{tikz}
\usetikzlibrary{positioning,arrows.meta,calc}

\newcommand{\NumTabOneAMathFiveHundredBaseNt}{84.7}% Avg@4 | cap-std 38912 shared base, CAP_STANDARDIZATION_RERUN_RESULTS.md (base_math500.json)

\newcommand{\NumTabOneAAimeTwentyFourBaseNt}{25.0}% Avg@12 | cap-std 32768, CAP_STANDARDIZATION_RERUN_RESULTS.md (B_base_aime24, jobs 18370-18372)

\newcommand{\NumTabOneAAimeTwentyFiveBaseNt}{19.7}% Avg@12 | cap-std 32768, CAP_STANDARDIZATION_RERUN_RESULTS.md (B_base_aime25, jobs 18373-18375)

\newcommand{\NumTabOneAHmmtTwentyFiveBaseNt}{13.1}% Avg@12 | cap-std 32768, CAP_STANDARDIZATION_RERUN_RESULTS.md (B_base_hmmt25, jobs 18376-18378)

\newcommand{\AvgTabOneABaseNtAvg}{35.6}% unweighted mean across displayed benchmarks

\newcommand{\NumTabOneAMathFiveHundredNtCanonical}{91.5}% Avg@4 | cap-std 38912, CAP_STANDARDIZATION_RERUN_RESULTS.md (A1 opsd_numina_8b/ck50, job 18123)

\newcommand{\NumTabOneAAimeTwentyFourNtCanonical}{66.7}% Avg@12 | cap-std 32768, CAP_STANDARDIZATION_RERUN_RESULTS.md (B_canon_aime24 opsd_numina_8b/ck50, jobs 18136-18138)

\newcommand{\NumTabOneAAimeTwentyFiveNtCanonical}{49.2}% Avg@12 | cap-std 32768, CAP_STANDARDIZATION_RERUN_RESULTS.md (B_canon_aime25, jobs 18139-18141)

\newcommand{\NumTabOneAHmmtTwentyFiveNtCanonical}{30.6}% Avg@12 | cap-std 32768, CAP_STANDARDIZATION_RERUN_RESULTS.md (B_canon_hmmt25, jobs 18142-18144)

\newcommand{\AvgTabOneANtCanonicalAvg}{59.5}% unweighted mean across displayed benchmarks

\newcommand{\NumTabOneAMathFiveHundredNtWrongRef}{94.4}% Avg@4 | results_wrongref.md:55

\newcommand{\NumTabOneAAimeTwentyFourNtWrongRef}{71.9}% Avg@12 | results_wrongref.md:76

\newcommand{\NumTabOneAAimeTwentyFiveNtWrongRef}{61.7}% Avg@12 | results_wrongref.md:76

\newcommand{\NumTabOneAHmmtTwentyFiveNtWrongRef}{42.2}% Avg@12 | results_wrongref.md:76

\newcommand{\AvgTabOneANtWrongRefAvg}{67.6}% unweighted mean across displayed benchmarks

\newcommand{\NumTabOneAMathFiveHundredThinkWithRef}{88.0}% Avg@4 | cap-std 38912, CAP_STANDARDIZATION_RERUN_RESULTS.md (A2 opsd_numina_tmon_g1024/ck200, job 18124)

\newcommand{\NumTabOneAAimeTwentyFourThinkWithRef}{65.3}% Avg@12 | pooled 3-shard val12 rep1.0 non-thinking eval, opsd_numina_tmon_g1024/checkpoint-200, 30 problems

\newcommand{\NumTabOneAAimeTwentyFiveThinkWithRef}{52.8}% Avg@12 | pooled 3-shard val12 rep1.0 non-thinking eval, opsd_numina_tmon_g1024/checkpoint-200, 30 problems

\newcommand{\NumTabOneAHmmtTwentyFiveThinkWithRef}{31.7}% Avg@12 | pooled 3-shard val12 rep1.0 non-thinking eval, opsd_numina_tmon_g1024/checkpoint-200, 30 problems

\newcommand{\AvgTabOneAThinkWithRefAvg}{59.4}% unweighted mean across displayed benchmarks

\newcommand{\NumTabOneAMathFiveHundredThinkNoRef}{89.5}% Avg@4 | cap-std 38912, CAP_STANDARDIZATION_RERUN_RESULTS.md (A3 opsd_numina_tmon_nopi/ck200, job 18125)

\newcommand{\NumTabOneAAimeTwentyFourThinkNoRef}{63.6}% Avg@12 | pooled 3-shard val12 rep1.0 non-thinking eval, opsd_numina_tmon_nopi/checkpoint-200, 30 problems

\newcommand{\NumTabOneAAimeTwentyFiveThinkNoRef}{51.1}% Avg@12 | pooled 3-shard val12 rep1.0 non-thinking eval, opsd_numina_tmon_nopi/checkpoint-200, 30 problems

\newcommand{\NumTabOneAHmmtTwentyFiveThinkNoRef}{36.9}% Avg@12 | pooled 3-shard val12 rep1.0 non-thinking eval, opsd_numina_tmon_nopi/checkpoint-200, 30 problems

\newcommand{\AvgTabOneAThinkNoRefAvg}{60.3}% unweighted mean across displayed benchmarks

\newcommand{\NumTabOneBMathFiveHundredBaseNt}{84.7}% Avg@4 | megascience/results_run1.md:33

\newcommand{\NumTabOneBGpqaBaseNt}{47.9}% Avg@12 | megascience/results_run1.md:325

\newcommand{\NumTabOneBMmluProBaseNt}{66.1}% Avg@1 | megascience/results_run1.md:132

\newcommand{\NumTabOneBUgPhysicsBaseNt}{12.6}% Avg@1 | megascience/results_run1.md:132

\newcommand{\AvgTabOneBBaseNtAvg}{52.8}% unweighted mean across displayed benchmarks

\newcommand{\NumTabOneBMathFiveHundredNtCanonical}{86.2}% Avg@4 | megascience/results_run1.md:35

\newcommand{\NumTabOneBGpqaNtCanonical}{49.7}% Avg@12 | unconditional Avg@12 (avg_at_n, unparseable counted incorrect), pooled 8-shard eval mega_opsd_canon/checkpoint-150 GPQA val_12 rep 1.3 over 198 problems, parseable 99.83 pct; conditional cond_acc was 49.75

\newcommand{\NumTabOneBMmluProNtCanonical}{64.0}% Avg@1 | megascience/results_run1.md:35

\newcommand{\NumTabOneBUgPhysicsNtCanonical}{13.4}% Avg@1 | megascience/results_run1.md:35

\newcommand{\AvgTabOneBNtCanonicalAvg}{53.3}% unweighted mean across displayed benchmarks

\newcommand{\NumTabOneBMathFiveHundredNtWrongRef}{88.0}% Avg@4 | cap-std 38912, CAP_STANDARDIZATION_RERUN_RESULTS.md (A4 mega_opsd_shufref/ck150, job 18126)

\newcommand{\NumTabOneBGpqaNtWrongRef}{49.8}% Avg@12 | unconditional Avg@12, pooled 3-shard val12 rep1.3 eval, mega_opsd_shufref/checkpoint-150 (within-subject deranged reference), 198 problems

\newcommand{\NumTabOneBMmluProNtWrongRef}{64.3}% Avg@1 | pooled 6-shard val1 rep1.3 eval, mega_opsd_shufref/checkpoint-150

\newcommand{\NumTabOneBUgPhysicsNtWrongRef}{12.7}% Avg@1 | pooled 3-shard val1 rep1.3 eval, mega_opsd_shufref/checkpoint-150

\newcommand{\AvgTabOneBNtWrongRefAvg}{53.7}% unweighted mean across displayed benchmarks

\newcommand{\NumTabOneBMathFiveHundredThinkWithRef}{92.8}% Avg@4 | megascience/results_no_pi.md:55

\newcommand{\NumTabOneBGpqaThinkWithRef}{53.4}% Avg@12 | megascience/results_no_pi.md:54 / results_run1.md:326

\newcommand{\NumTabOneBMmluProThinkWithRef}{69.5}% Avg@1 | megascience/results_no_pi.md:56

\newcommand{\NumTabOneBUgPhysicsThinkWithRef}{27.1}% Avg@1 | megascience/results_no_pi.md:57

\newcommand{\AvgTabOneBThinkWithRefAvg}{60.7}% unweighted mean across displayed benchmarks

\newcommand{\NumTabOneBMathFiveHundredThinkNoRef}{92.7}% Avg@4 | megascience/results_no_pi.md:55

\newcommand{\NumTabOneBGpqaThinkNoRef}{53.5}% Avg@12 | megascience/results_no_pi.md:85

\newcommand{\NumTabOneBMmluProThinkNoRef}{69.8}% Avg@1 | megascience/results_no_pi.md:56

\newcommand{\NumTabOneBUgPhysicsThinkNoRef}{27.2}% Avg@1 | megascience/results_no_pi.md:57

\newcommand{\AvgTabOneBThinkNoRefAvg}{60.8}% unweighted mean across displayed benchmarks

\title{Rethinking Privileged Information \\
in On-Policy Self-Distillation}

\author{
Samyak Shrestha \\
FirstPrinciples \\
\texttt{samyak.shrestha@utdallas.edu}
\And
Alexander Tessier \\
FirstPrinciples \\
\texttt{alexander@firstprinciples.com}
}


\begin{document}
\raggedbottom
\maketitle
\pagestyle{plain}

\begin{abstract}
On-policy self-distillation (OPSD) trains a student on its own responses using
token-level supervision from the same model conditioned on privileged reference
information. We investigate whether performance gains from OPSD show that the student
learned the information in the reference or instead reflect recovery of reasoning
behavior already present in the base model. We perform OPSD experiments on science and
mathematics datasets using Qwen3 models ranging from 1.7B to 8B. Our analysis framework
separates the supervision induced by the reference from the supervision provided by the
teacher without the reference and measures how each aligns with changes in the
student's predictions. The correct reference does not provide a consistent performance
benefit across teacher generation modes, model sizes, and training datasets. Students
can improve without the correct reference, and a solution from another problem can
outperform the correct solution on several mathematical reasoning benchmarks. The
student's predictions align more strongly with the base model's thinking behavior than
with the supervision induced by the reference, but controls constructed from other
problems reproduce much of both alignments.
Moreover, stronger alignment attributable to the correct reference does not reliably
coincide with a greater performance benefit from the reference. Performance gains and
distributional alignment alone therefore cannot determine how privileged reference
information contributes to student learning in OPSD.
\end{abstract}

\section{Introduction}
\label{sec:intro}

On-policy distillation (OPD) has become a widely used method in the post-training of
large language models (LLMs), including recent releases such as
DeepSeek-V4~\citep{deepseekv4}, GLM-5~\citep{glm5},
Kimi K3~\citep{kimik3}, Nemotron-Cascade 2~\citep{nemotroncascade2}, and
Qwen3~\citep{qwen3}. Unlike supervised fine-tuning (SFT), which is prone to exposure
bias from training on fixed target sequences~\citep{bengio2015scheduled}, OPD trains the
student on trajectories sampled from its current policy. A teacher provides dense,
token-level supervision on the states visited by the
student~\citep{agarwal2024gkd,gu2024minillm,lu2025onpolicy}. On-policy
self-distillation (OPSD) uses the same model as both teacher and student under different
contexts~\citep{zhao2026opsd}. The student generates a response from the problem alone,
while the teacher evaluates the same response with access to privileged information,
typically a reference solution unavailable to the student.

The central premise of OPSD is that privileged reference information improves the
teacher's supervision and transfers useful information to the
student~\citep{zhao2026opsd}. However, increasing the teacher's exposure to a reference
is not consistently beneficial and can alter uncertainty expression or reasoning
behavior~\citep{han2026adaptive,kim2026degrade,kaur2026rethinking}. Reference-conditioned
supervision may also contain information that the student cannot use when the reference
is absent at inference time~\citep{zhu2026manyfaces,shen2026purified}. Performance alone
therefore cannot determine whether the student learned from the reference. This question
becomes more difficult when the teacher operates in thinking mode because the teacher
then differs from the student in both its context and its generation mode.

We therefore ask three questions. First, is the correct reference necessary for OPSD to
improve performance? Second, must the reference correspond to the problem being solved?
Third, do the gains reflect learning from the reference or recovery of reasoning behavior
already present in the base model? We study these questions using Qwen3-8B, Qwen3-4B,
and Qwen3-1.7B students trained on NuminaMath and MegaScience. For OPSD, pairing a
non-thinking student with a thinking teacher produces the largest performance gains
among the tested mode combinations~\citep{zhao2026opsd}. In our experiments, the student
remains in non-thinking mode during both training and evaluation, while we vary the
teacher's generation mode and reference context. Holding the student's generation mode
fixed isolates these teacher-side effects. It also allows us to test whether supervision
from a thinking teacher moves the student toward behavior associated with the base
model's thinking mode.

We introduce an analysis framework that separates the supervision induced by the
reference from the remaining teacher supervision. We measure how each form of
supervision aligns with changes in the student's predictions during training and repeat
the measurements using teacher contexts from other problems.
Figure~\ref{fig:setup} summarizes the OPSD setup and the teacher and reference conditions
used in our experiments.

% Fig 1 --- the OPSD setup schematic. Hand-authored TikZ (no data source).
% The on-policy self-distillation loop plus the advice-or-recovery question. Requires, in the
% preamble: \usepackage{tikz} and \usetikzlibrary{positioning,arrows.meta,calc}.
% Reserved color: vermillion = reference (matches D_ref everywhere).
\definecolor{okverm}{HTML}{D55E00}
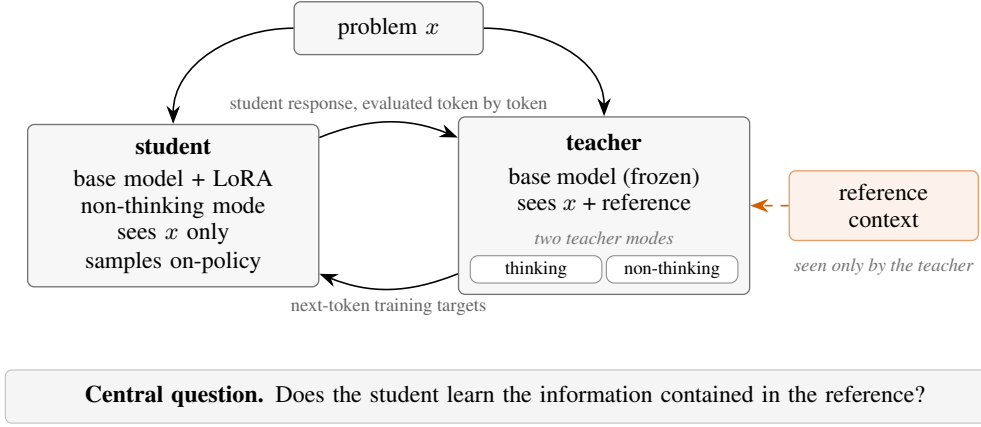
\begin{figure}[t]
\centering
\begin{tikzpicture}[
  font=\footnotesize\fontfamily{ptm}\selectfont,
  >={Stealth[length=2.4mm]},
  box/.style={draw=black!55, rounded corners=2.5pt, line width=0.5pt,
              align=center, inner sep=5pt, fill=black!3},
  small/.style={box, text width=2.15cm, minimum height=0.7cm},
  big/.style={box, text width=3.5cm, minimum height=2.05cm},
  bigA/.style={box, text width=3.5cm, minimum height=2.35cm},
  ref/.style={small, fill=okverm!8, draw=okverm!60},
  pill/.style={draw=black!45, rounded corners=3pt, fill=white,
               inner sep=2pt, text width=1.6cm, align=center,
               font=\scriptsize\fontfamily{ptm}\selectfont},
  lab/.style={font=\scriptsize\fontfamily{ptm}\selectfont,
              text=black!60, align=center},
  ilab/.style={font=\scriptsize\usefont{T1}{ptm}{m}{it},
               text=black!55, align=center},
  flow/.style={->, line width=0.6pt},
]
% everything but the banner, captured in a bounding box so the banner can center under it
\begin{scope}[local bounding box=diag]
% student (left), teacher (right, with the two-mode toggle inside)
\node[big] (S) at (3.0,0)
  {{\usefont{T1}{ptm}{b}{n} student}\\[2pt] base model + LoRA\\ non-thinking mode\\
   sees $x$ only\\[1pt] samples on-policy};
\node[bigA] (T) at (8.7,0) {};
\node[anchor=north, align=center] at ([yshift=-3pt]T.north)
  {{\usefont{T1}{ptm}{b}{n} teacher}};
\node[align=center] at ([yshift=0.22cm]T.center)
  {base model (frozen)\\ sees $x$ + reference};
\node[ilab] at ([yshift=0.76cm]T.south) {two teacher modes};
\node[pill] (pT) at ([xshift=-0.90cm,yshift=0.34cm]T.south) {thinking};
\node[pill] (pN) at ([xshift= 0.90cm,yshift=0.34cm]T.south) {non-thinking};
% problem (top, feeds both), reference (right, teacher only)
\node[small] (P) at (5.85,2.35) {problem $x$};
\node[ref] (R) at (12.4,0) {reference context};
\node[ilab, below=1mm of R] {seen only by the teacher};
\draw[flow] (P.west) to[out=180,in=90] (S.north);
\draw[flow] (P.east) to[out=0,in=90] (T.north);
\draw[flow, okverm, dashed] (R.west) -- (T.east);
% on-policy loop
\draw[flow] (S.25) to[bend left=22]
  node[above,lab]{student response, evaluated token by token} (T.155);
\draw[flow] (T.205) to[bend left=22]
  node[below,lab]{next-token training targets} (S.-25);
\end{scope}
% question banner, centered under the diagram bounding box
\node[draw=black!25, rounded corners=2.5pt, fill=black!4, inner sep=6pt,
      text width=12.8cm, align=center, anchor=north] (Q) at ([yshift=-6mm]diag.south)
  {{\usefont{T1}{ptm}{b}{n} Central question.}
   Does the student learn the information contained in the reference?};
\end{tikzpicture}
\caption{OPSD setup and experimental conditions. The student remains in non-thinking
mode while we vary the teacher's generation mode and reference context.}
\label{fig:setup}
\end{figure}

Our contributions are as follows:

\begin{itemize}
    \item We show that the benefit of the correct reference depends on the teacher's
    generation mode, the model size, and the training dataset. Students can improve
    without the correct reference, and a solution from another problem can outperform
    the correct solution on several mathematical reasoning benchmarks.

    \item We introduce an analysis framework that separates reference-induced
    supervision from the remaining teacher supervision and measures how each aligns
    with changes in the student's predictions. The change in the student's predictions
    aligns more strongly with the base model's thinking behavior than with the
    supervision induced by the reference.

    \item We compare these alignments with controls constructed from other problems.
    The controls reproduce much of the apparent alignment, and stronger alignment
    attributable to the correct reference does not reliably coincide with a greater
    performance benefit from the reference. Together, these results show that
    performance gains and distributional alignment alone cannot determine how
    privileged reference information contributes to student learning in OPSD.
\end{itemize}

\section{Related work}
\label{sec:related}

\paragraph{On-policy distillation.}
Knowledge distillation trains a student to match the output distribution of a teacher
model~\citep{hinton2015distilling}. Sequence-level distillation adapts this approach to
autoregressive models by training the student on complete responses generated by the
teacher~\citep{kim2016sequence}. Since these responses are not generated by the student's current policy, the prefixes
seen during training may differ from those encountered during inference, creating
exposure bias and allowing errors to accumulate throughout the generated
response~\citep{bengio2015scheduled,agarwal2024gkd}.
Imitation-based distillation addresses this mismatch by training on student-generated
trajectories, with the teacher providing supervision at the states visited by the
student~\citep{lin2020autoregressive}. OPD applies this principle to LLMs by evaluating the teacher on trajectories sampled
from the student's current policy~\citep{gu2024minillm,lu2025onpolicy}. Training
minimizes a token-level KL divergence between the teacher's and student's next-token
distributions, providing dense supervision at every position in the generated
response~\citep{agarwal2024gkd}.

\paragraph{On-policy self-distillation.}
On-policy self-distillation (OPSD) removes the need for a separate teacher by
instantiating the teacher and student from a single model under different
contexts~\citep{zhao2026opsd}. The student is conditioned on the problem alone, while
the teacher also receives privileged information such as a reference solution. The
student generates the training trajectories, and the teacher provides token-level
supervision by evaluating the same prefixes with access to the additional context.
Closely related work applies this principle using expert demonstrations and textual
feedback~\citep{shenfeld2026sdft,hubotter2026sdpo}. Other methods extend the teacher's
context beyond reference solutions to include experiential knowledge extracted from
historical solution traces and optimized system prompts~\citep{ye2026opcd}, action-only
privileged information~\citep{penaloza2026privileged}, source
documents~\citep{stein2026gates}, and natural-language skills extracted from completed
trajectories~\citep{wang2026skillsd}.
Together, these methods use contextual information available during training to create a
more informed teacher policy.

\paragraph{Privileged context in OPSD.}
A more informed teacher does not necessarily provide more useful supervision. Privileged
information specific to an individual problem may not transfer reliably to a student
that lacks that information at inference time~\citep{zhu2026manyfaces}. Providing more
of the reference can increase the mismatch between teacher and
student~\citep{han2026adaptive}, while rich teacher context can suppress uncertainty and
impair out-of-distribution generalization~\citep{kim2026degrade}. Privileged-context
distillation can also degrade thinking models by suppressing verification and
self-correction~\citep{kaur2026rethinking}. Proposed remedies adapt the amount of
reference shown to the teacher~\citep{han2026adaptive}, reduce the influence of
high-entropy token positions~\citep{ke2026egrsd}, restrict distillation to identified
reasoning errors~\citep{zhao2026rosd}, or remove reference-specific shortcuts from the
teacher's signal~\citep{shen2026purified}. Concurrent work finds that a solution from
another problem can preserve or improve OPSD
performance~\citep{ichihara2026op2sd} and shows that likelihood changes induced by
privileged context do not necessarily provide useful token
credit~\citep{nguyen2026privilegedlikelihood}. Our study complements this work by
separating reference-dependent from reference-free teacher supervision and comparing
how each relates to changes in the student's predictions using controls constructed
from other problems.

\section{Analysis Framework}
\label{sec:setup}

\subsection{Training Objective}
\label{sec:setup:prelim}

In OPSD, a student model $\pi_S$ generates trajectories from its current policy, while a
frozen teacher $\pi_T$ instantiated from the same base model provides token-level
supervision with access to a reference solution~\citep{agarwal2024gkd,zhao2026opsd}.
Let $\mathcal{D}$ denote the training dataset. For a problem $x\sim\mathcal{D}$ with
reference solution $r$, the student samples a trajectory
$y\sim\pi_S(\cdot\mid x)$. The training objective is
\begin{equation}
\label{eq:opsd}
\mathcal{L}(\pi_S)
=
\mathbb{E}_{x\sim\mathcal{D},\;y\sim\pi_S(\cdot\mid x)}
\left[
\sum_{t=1}^{|y|}
D_{\mathrm{JSD}}\!\left(
\pi_T(\cdot\mid x,r,y_{<t})
\,\Vert\,
\pi_S(\cdot\mid x,y_{<t})
\right)
\right],
\end{equation}
where $D_{\mathrm{JSD}}$ denotes the generalized Jensen--Shannon divergence.

\subsection{Reference-Dependent and Reference-Free Supervision}
\label{sec:setup:decomp}

Equation~\ref{eq:opsd} specifies how the student learns from the teacher, but it does not
identify which part of the teacher's supervision comes from the reference. Motivated by
the decomposition approach introduced in Purified OPSD~\citep{shen2026purified}, we
isolate the contribution made by the reference by comparing the teacher's next-token
distributions with and without it.

To make this comparison, we let the student at checkpoint $c$ generate a response
$y^{(c)}\sim\pi_S^{(c)}(\cdot\mid x)$ to problem $x$. At each position $t$ of the
response, let $S_c$ denote the student's next-token distribution. Let $T$ and $N$ denote
the next-token distributions of the teacher with and without reference $r$,
respectively. Thus, for every token $v$ in the vocabulary $\mathcal{V}$,
\begin{equation}
\label{eq:distributions}
T(v)
=
\pi_T(v\mid x,r,y^{(c)}_{<t}),
\qquad
N(v)
=
\pi_T(v\mid x,y^{(c)}_{<t}),
\qquad
S_c(v)
=
\pi_S^{(c)}(v\mid x,y^{(c)}_{<t}).
\end{equation}

We represent each distribution by its vector of log probabilities:
\[
\log T
:=
\bigl(\log T(v)\bigr)_{v\in\mathcal{V}}
\in \mathbb{R}^{|\mathcal{V}|},
\]
with $\log N$ and $\log S_c$ defined analogously. The total difference
$D_{\mathrm{tot}}^{(c)}$ between the log-probability vectors of the teacher and the
student can be decomposed as
\begin{equation}
\label{eq:decomposition}
\underbrace{\log T-\log S_c}_{D_{\mathrm{tot}}^{(c)}}
=
\underbrace{\log T-\log N}_{D_{\mathrm{ref}}^{(c)}}
+
\underbrace{\log N-\log S_c}_{D_{\mathrm{rec}}^{(c)}},
\end{equation}
where $D_{\mathrm{ref}}^{(c)}$ is the reference direction that measures the difference
between the teacher with and without the reference, and $D_{\mathrm{rec}}^{(c)}$ is the
recovery direction that measures the difference between the teacher without the
reference and the student.

\subsection{Composition of Teacher Supervision}
\label{sec:setup:composition}

Before analyzing how the student changes during training, we measure the component of
the reference direction $D_{\mathrm{ref}}$ along the total teacher supervision
$D_{\mathrm{tot}}$. We compute this measurement at training step 0 using responses
generated by the initial student.

For each problem $x$, the initial student generates a response
$y^{(0)}\sim\pi_S^{(0)}(\cdot\mid x)$. Let $\mathcal{I}_0$ denote the set of all
problem-position pairs $(x,t)$ in these responses. For any two directions $A$ and $B$,
let
$A^{x,t},B^{x,t}\in\mathbb{R}^{|\mathcal{V}|}$ denote their values at position $t$ of
the response to problem $x$. We define the inner product and its induced norm as
\begin{equation}
\label{eq:pooled}
\langle A,B\rangle_{\mathcal{I}_0}
=
\frac{1}{|\mathcal{I}_0|}
\sum_{(x,t)\in\mathcal{I}_0}
\langle A^{x,t},B^{x,t}\rangle,
\qquad
\lVert A\rVert_{\mathcal{I}_0}
=
\sqrt{\langle A,A\rangle_{\mathcal{I}_0}},
\end{equation}
where $\lVert A\rVert_{\mathcal{I}_0}$ is the root-mean-square magnitude of $A$ across
problem-position pairs.

We define the reference projection coefficient as
\begin{equation}
\label{eq:shares}
w_{\mathrm{ref}}
=
\frac{
\left\langle D_{\mathrm{ref}},D_{\mathrm{tot}}\right\rangle_{\mathcal{I}_0}
}{
\left\lVert D_{\mathrm{tot}}\right\rVert_{\mathcal{I}_0}^{2}
}.
\end{equation}

\subsection{Changes in Student Predictions on Fixed Responses}
\label{sec:setup:fixed-text}

Our analysis so far has focused on the teacher's supervision. To measure how the
student's predictions change during training, we evaluate every checkpoint $c$ on the
responses generated by the initial student at training step 0.

At each position $t$, let $S_0$ and $S_c$ denote
the next-token distributions of the initial student and the student at checkpoint $c$,
conditioned on the same generated prefix:
\begin{equation}
\label{eq:fixed-student-distributions}
S_0(v)
=
\pi_S^{(0)}(v\mid x,y^{(0)}_{<t}),
\qquad
S_c(v)
=
\pi_S^{(c)}(v\mid x,y^{(0)}_{<t}).
\end{equation}
We also compute the teacher distributions $T$ and $N$ from
Section~\ref{sec:setup:decomp} on these prefixes. Let $B$ denote the distribution
produced by the base model in thinking mode without the reference:
\[
B(v)
=
\pi_T^{\mathrm{think}}(v\mid x,y^{(0)}_{<t}).
\]

We define the change in the student's log probabilities as
\begin{equation}
\label{eq:student-change}
\Delta S_c
=
\log S_c-\log S_0.
\end{equation}
The teacher directions are defined on the same fixed responses:
\[
D_{\mathrm{ref}}
=
\log T-\log N,
\qquad
D_{\mathrm{rec}}
=
\log N-\log S_0,
\qquad
D_{\mathrm{think}}
=
\log B-\log S_0.
\]
The responses, generated prefixes, and teacher distributions remain fixed across
checkpoints. Only the student distribution $S_c$ and the resulting change
$\Delta S_c$ vary with $c$. This allows us to compare changes in the student's
predictions while keeping the text on which the distributions are computed fixed.

When the teacher operates in non-thinking mode without the reference, its model and
context are the same as those of the initial student. Therefore, $N=S_0$ and
$D_{\mathrm{rec}}=0$.

\subsection{Relationship Between Teacher Supervision and Student Predictions}
\label{sec:setup:student-metrics}

We characterize the student's change in three ways. We first measure its projection
onto each teacher direction and its cosine similarity with that direction. We then
measure how much of the student's change can be represented by a linear combination of
two teacher directions. Finally, we measure the magnitude of $\Delta S_c$ and the KL
divergence from the reference-free teacher distribution $N$ to $S_c$. Let $D$ denote
any nonzero teacher direction defined in Section~\ref{sec:setup:fixed-text}.

We measure the component of $\Delta S_c$ along direction $D$ using the projection
coefficient
\begin{equation}
\label{eq:student-projection}
\beta_D^{(c)}
=
\frac{
\langle\Delta S_c,D\rangle_{\mathcal{I}_0}
}{
\lVert D\rVert_{\mathcal{I}_0}^{2}
}.
\end{equation}
The projection coefficient depends on the relative magnitudes of $\Delta S_c$ and $D$
as well as on their directions. To compare the directions independently of magnitude,
we use cosine similarity:
\begin{equation}
\label{eq:student-alignment}
\cos\theta_D^{(c)}
=
\frac{
\langle\Delta S_c,D\rangle_{\mathcal{I}_0}
}{
\lVert\Delta S_c\rVert_{\mathcal{I}_0}
\lVert D\rVert_{\mathcal{I}_0}
}.
\end{equation}
When comparing teacher directions, we compute the cosine similarity between the
reference direction $D_{\mathrm{ref}}$ and the thinking direction
$D_{\mathrm{think}}$.

We next measure the fraction of
$\lVert\Delta S_c\rVert_{\mathcal{I}_0}^{2}$ preserved by its orthogonal projection
onto the span of two teacher directions. We use $D_{\mathrm{ref}}$ and
$D_{\mathrm{rec}}$ for a thinking teacher. Because $D_{\mathrm{rec}}=0$ for a
non-thinking teacher, we use $D_{\mathrm{ref}}$ and $D_{\mathrm{think}}$ instead. At
each problem-position pair, we define the span as
\begin{equation}
\label{eq:teacher-span}
\mathcal W^{x,t}
=
\begin{cases}
\operatorname{span}
\left\{
D_{\mathrm{ref}}^{x,t},
D_{\mathrm{rec}}^{x,t}
\right\},
& \text{thinking teacher},
\\[3pt]
\operatorname{span}
\left\{
D_{\mathrm{ref}}^{x,t},
D_{\mathrm{think}}^{x,t}
\right\},
& \text{non-thinking teacher}.
\end{cases}
\end{equation}
Let $P_{\mathcal W^{x,t}}$ denote the orthogonal projection onto this span. We define
this fraction as
\begin{equation}
\label{eq:student-span}
R_c^2
=
\frac{
\sum_{(x,t)\in\mathcal{I}_0}
\left\lVert
P_{\mathcal W^{x,t}}\Delta S_c^{x,t}
\right\rVert_2^2
}{
\sum_{(x,t)\in\mathcal{I}_0}
\left\lVert
\Delta S_c^{x,t}
\right\rVert_2^2
}.
\end{equation}

Finally, we measure the magnitude of $\Delta S_c$ using
$\lVert\Delta S_c\rVert_{\mathcal{I}_0}$. We measure the average token-level KL
divergence from $N$ to $S_c$ as
\[
\operatorname{KL}_{\mathcal{I}_0}(N\Vert S_c)
=
\frac{1}{|\mathcal{I}_0|}
\sum_{(x,t)\in\mathcal{I}_0}
\operatorname{KL}\!\left(N^{x,t}\Vert S_c^{x,t}\right).
\]

\subsection{Problem-Specific Effects of Teacher Supervision}
\label{sec:setup:controls}

The metrics in Section~\ref{sec:setup:student-metrics} compare the student's change with
the teacher directions. Because the teacher and student share the same base model and
evaluate the same generated prefix, some of the measured similarity may not be specific
to the problem. We control for this by repeating each comparison using the teacher
context from a different problem.

To construct the control, we keep the student-generated response fixed and condition
the teacher on a different problem. If the teacher receives a reference, it receives
the reference for this different problem. This produces the control distributions
$T^*$, $N^*$, and $B^*$. The control directions are:
\[
D_{\mathrm{ref}}^*
=
\log T^*-\log N^*,
\qquad
D_{\mathrm{rec}}^*
=
\log N^*-\log S_0,
\qquad
D_{\mathrm{think}}^*
=
\log B^*-\log S_0.
\]
The student distributions $S_0$ and $S_c$ remain fixed, so $\Delta S_c$ does not
change. When comparing two teacher directions, we keep the problem and generated prefix
fixed and replace only the reference with one from another problem.
This comparison measures whether the alignment between the two teacher directions
depends on using the correct reference.

For the measurements defined in Section~\ref{sec:setup:student-metrics}, we replace the
original teacher directions with the control directions defined above. The difference
between the original and control measurements isolates dependence on the correct problem
context. Because assigning a different teacher context can change a direction's
magnitude, we use cosine similarity as the primary directional measure and report
projection coefficients with the corresponding direction norms. We estimate uncertainty
using a paired bootstrap over problems. To verify that the results do not depend on a
particular pairing between evaluated and control problems, we repeat each control
construction under three different pairings while keeping the student responses and
original measurements fixed.

\section{Experiments}
\label{sec:results}
\subsection{Experimental Setup}
\label{sec:experiments:setup}

\paragraph{Models and datasets.}
Our main experiments use Qwen3-8B~\citep{qwen3}. We repeat the core performance and
distributional comparisons with Qwen3-4B and Qwen3-1.7B. Complete results for these
models appear in Appendix~\ref{app:scale}. At each scale, the student and teacher are
instantiated from the same base model. Each student uses LoRA with rank
64 and alpha 128~\citep{hu2022lora} and operates in non-thinking mode. The teacher is
frozen and operates in either thinking or non-thinking mode. We curate a
10,000-problem subset of NuminaMath~\citep{numinamath} and another 10,000-problem
subset of MegaScience~\citep{fan2025megascience}. Both subsets contain worked solutions
with extractable boxed answers, and we remove overlaps with the corresponding
evaluation benchmarks.

\paragraph{Training conditions and implementation.}
At all three scales, we train students using thinking and non-thinking teachers. For
the thinking teacher, we compare training with and without the reference. For the
non-thinking teacher, we compare the correct reference with a reference from another
problem. At 8B, we additionally compare answer-only references, abstract hints,
canonical solutions, and reasoning traces generated by Qwen3-8B and Qwen3-32B. We use
the same 4,953 NuminaMath problems for the answer-only references, canonical solutions,
and reasoning traces. The abstract-hint condition uses 4,898 of these problems after
removing 55 hints that reveal the answer. We train for 300 steps with a learning rate of
$5\times10^{-6}$ and an effective batch size of 32. We generate on-policy student
trajectories with vLLM~\citep{kwon2023vllm} at temperature 1.1 with top-$p$ 0.95 and
top-$k$ 20. We set the generalized Jensen--Shannon interpolation parameter to $\beta=0$.
For experiments with a non-thinking teacher, we set the maximum student completion
length to 4,096 tokens on NuminaMath and 2,048 tokens on MegaScience, with pointwise
clipping at $10^{-7}$. For experiments with a thinking teacher, we set the maximum
student completion length to 1,024 tokens and use pointwise clipping at 0.06.

\paragraph{Evaluation.}
We evaluate mathematical reasoning on MATH-500~\citep{hendrycks2021math}, AIME 2024,
AIME 2025, and HMMT 2025, and scientific reasoning on
GPQA-Diamond~\citep{rein2023gpqa}, MMLU-Pro~\citep{wang2024mmlupro}, and
UGPhysics~\citep{xu2025ugphysics}. We report Avg@$k$, the mean accuracy across $k$
sampled responses. The
main AIME, HMMT, and GPQA-Diamond comparisons use $k=12$. MATH-500 uses $k=4$,
while the 8B reference-type comparison in
Table~\ref{tab:appendix-reference-type-performance} uses $k=12$.
MMLU-Pro and UGPhysics use $k=1$. The primary evaluations use non-thinking mode to hold
the student's generation mode fixed. Appendix~\ref{app:thinking-eval} reports an
auxiliary Qwen3-8B evaluation with thinking enabled. All evaluations use temperature
1.0, with top-$p$ 0.8 in non-thinking mode and 0.95 in thinking mode. We use a
repetition penalty of 1.3 for students trained with a non-thinking teacher and 1.0 for
students trained with a thinking teacher. We grade free-response mathematics by
symbolic equivalence and use
benchmark-specific answer matching for the remaining evaluations. Uncertainty
estimation, checkpoint-selection procedures, and complete evaluation settings are
reported in Appendices~\ref{app:evaluation} and~\ref{app:scale}. We describe one
student as performing better than another only when the paired 95\% interval for their
score difference does not contain zero.

\subsection{Effects of Teacher Mode, Reference Information, and Model Scale on Performance}
\label{sec:results:outcomes}

Table~\ref{tab:main} reports the complete Qwen3-8B results. On NuminaMath, students
trained with either teacher mode score above the non-thinking base model on average. On
MegaScience, students trained with a thinking teacher score above the non-thinking base
model on all four benchmarks, whereas students trained with a non-thinking teacher show
smaller and less consistent differences. Adding the reference to a thinking teacher
provides no consistent advantage at 8B. On NuminaMath, the student trained with the
reference scores higher on AIME 2024 and AIME 2025, while the student trained without
it scores higher on MATH-500 and HMMT 2025. On MegaScience, their scores differ by at
most 0.3 points across the four benchmarks.

% Table 1 --- main benchmark results. Values are generated in numbers.tex.
% Presentation-only: single self-describing Method column, no numerical bolding.
\begin{table}[t]
\centering
\caption{Main benchmark results for Qwen3-8B. The base model and all OPSD
students are evaluated in non-thinking mode. Ref.\ denotes the reference
solution. Avg.\ is the unweighted mean across benchmarks. AIME, HMMT, and
GPQA-Diamond use Avg@12, MATH-500 uses Avg@4, and MMLU-Pro and UGPhysics use
Avg@1.}
\label{tab:main}
\vspace{7pt}
\footnotesize
\setlength{\tabcolsep}{1.5pt}
\renewcommand{\arraystretch}{1.04}
\begin{tabularx}{\linewidth}{@{}l
  >{\raggedleft\arraybackslash\hsize=1.174\hsize}X
  >{\raggedleft\arraybackslash\hsize=0.982\hsize}X
  >{\raggedleft\arraybackslash\hsize=1.220\hsize}X
  >{\raggedleft\arraybackslash\hsize=1.146\hsize}X
  >{\raggedleft\arraybackslash\hsize=0.478\hsize}X@{}}
\toprule
\textbf{Method} &
\multicolumn{5}{c}{\textbf{NuminaMath}} \\
\cmidrule(lr){2-6}
 &
\textbf{MATH-500} & \textbf{AIME24} & \textbf{AIME25} &
\textbf{HMMT25} & \textbf{Avg.} \\
\midrule
\rowcolor{gray!10}
Base model &
  \NumTabOneAMathFiveHundredBaseNt &
  \NumTabOneAAimeTwentyFourBaseNt &
  \NumTabOneAAimeTwentyFiveBaseNt &
  \NumTabOneAHmmtTwentyFiveBaseNt &
  \AvgTabOneABaseNtAvg \\
\addlinespace[2pt]
OPSD, non-thinking teacher, correct ref. &
  \NumTabOneAMathFiveHundredNtCanonical &
  \NumTabOneAAimeTwentyFourNtCanonical &
  \NumTabOneAAimeTwentyFiveNtCanonical &
  \NumTabOneAHmmtTwentyFiveNtCanonical &
  \AvgTabOneANtCanonicalAvg \\
OPSD, non-thinking teacher, mismatched ref. &
  \NumTabOneAMathFiveHundredNtWrongRef &
  \NumTabOneAAimeTwentyFourNtWrongRef &
  \NumTabOneAAimeTwentyFiveNtWrongRef &
  \NumTabOneAHmmtTwentyFiveNtWrongRef &
  \AvgTabOneANtWrongRefAvg \\
OPSD, thinking teacher, correct ref. &
  \NumTabOneAMathFiveHundredThinkWithRef &
  \NumTabOneAAimeTwentyFourThinkWithRef &
  \NumTabOneAAimeTwentyFiveThinkWithRef &
  \NumTabOneAHmmtTwentyFiveThinkWithRef &
  \AvgTabOneAThinkWithRefAvg \\
OPSD, thinking teacher, no ref. &
  \NumTabOneAMathFiveHundredThinkNoRef &
  \NumTabOneAAimeTwentyFourThinkNoRef &
  \NumTabOneAAimeTwentyFiveThinkNoRef &
  \NumTabOneAHmmtTwentyFiveThinkNoRef &
  \AvgTabOneAThinkNoRefAvg \\
\midrule
\textbf{Method} &
\multicolumn{5}{c}{\textbf{MegaScience}} \\
\cmidrule(lr){2-6}
 &
\textbf{MATH-500} & \textbf{GPQA-D} & \textbf{MMLU-Pro} &
\textbf{UGPhysics} & \textbf{Avg.} \\
\midrule
\rowcolor{gray!10}
Base model &
  \NumTabOneBMathFiveHundredBaseNt &
  \NumTabOneBGpqaBaseNt &
  \NumTabOneBMmluProBaseNt &
  \NumTabOneBUgPhysicsBaseNt &
  \AvgTabOneBBaseNtAvg \\
\addlinespace[2pt]
OPSD, non-thinking teacher, correct ref. &
  \NumTabOneBMathFiveHundredNtCanonical &
  \NumTabOneBGpqaNtCanonical &
  \NumTabOneBMmluProNtCanonical &
  \NumTabOneBUgPhysicsNtCanonical &
  \AvgTabOneBNtCanonicalAvg \\
OPSD, non-thinking teacher, mismatched ref. &
  \NumTabOneBMathFiveHundredNtWrongRef &
  \NumTabOneBGpqaNtWrongRef &
  \NumTabOneBMmluProNtWrongRef &
  \NumTabOneBUgPhysicsNtWrongRef &
  \AvgTabOneBNtWrongRefAvg \\
OPSD, thinking teacher, correct ref. &
  \NumTabOneBMathFiveHundredThinkWithRef &
  \NumTabOneBGpqaThinkWithRef &
  \NumTabOneBMmluProThinkWithRef &
  \NumTabOneBUgPhysicsThinkWithRef &
  \AvgTabOneBThinkWithRefAvg \\
OPSD, thinking teacher, no ref. &
  \NumTabOneBMathFiveHundredThinkNoRef &
  \NumTabOneBGpqaThinkNoRef &
  \NumTabOneBMmluProThinkNoRef &
  \NumTabOneBUgPhysicsThinkNoRef &
  \AvgTabOneBThinkNoRefAvg \\
\bottomrule
\end{tabularx}
\end{table}

At 4B and 1.7B, the effect of the reference depends on the training dataset. On
NuminaMath, paired comparisons support improvements on AIME 2024, AIME 2025, and HMMT
2025 at 4B, and on AIME 2024 and HMMT 2025 at 1.7B. For students trained on
MegaScience, the student trained with the reference does not outperform the student
trained without it at 4B. At 1.7B, removing the reference improves all three science
benchmarks. Complete results appear in
Tables~\ref{tab:appendix-4b-performance} and
\ref{tab:appendix-1p7b-performance}.

Under a non-thinking teacher, replacing the correct solution with a solution from
another problem improves AIME 2025 and HMMT 2025 at 8B, and all three competition
mathematics benchmarks at 4B and 1.7B. For students trained on MegaScience, the student
trained with a solution from another problem does not outperform the student trained
with the correct solution on any of the three science benchmarks at 4B or 1.7B. The
corresponding 8B scores and paired comparisons appear in
Table~\ref{tab:appendix-other-problem-performance}. Across the tested models,
the performance value of the reference therefore depends on the teacher's generation
mode, the model scale, and the training dataset.

\FloatBarrier

\subsection{Effects of Reference Information on the Teacher's Predictions}
\label{sec:results:signal}

Performance does not reveal how the reference changes the teacher's predictions. We
measure this effect at Qwen3-8B using the decomposition defined in
Section~\ref{sec:setup:decomp}. At training step 0, the reference projection coefficient
$w_{\mathrm{ref}}$ is 0.46 on NuminaMath and 0.39 on MegaScience. The reference
therefore changes the teacher's predictions even though adding it provides no consistent
performance advantage at this scale. On NuminaMath, under a non-thinking teacher, we
compare the reference direction $D_{\mathrm{ref}}$, induced by adding the reference, with the
thinking direction $D_{\mathrm{think}}$, induced by enabling thinking mode. We construct
the control $D_{\mathrm{ref}}^*$ by replacing the correct reference with a reference
from another problem.

\begin{figure}[!t]
\centering
\includegraphics[width=0.65\linewidth]{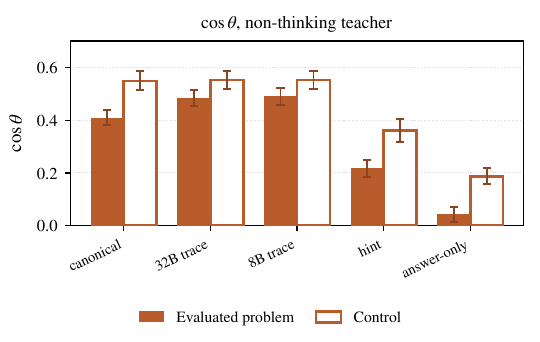}
\caption{Cosine similarity between the reference and thinking directions on NuminaMath
under a non-thinking teacher, across five reference types. Error bars show 1.96
bootstrap standard errors across problems.}
\label{fig:signal}
\end{figure}

Figure~\ref{fig:signal} reports this comparison across five reference types. In
every condition, the cosine similarity between $D_{\mathrm{ref}}$ and
$D_{\mathrm{think}}$ is lower than the cosine similarity between
$D_{\mathrm{ref}}^*$ and $D_{\mathrm{think}}$. This ordering holds under all three
control assignments. The direction induced by the correct reference therefore does not
align more closely with $D_{\mathrm{think}}$ than the direction induced by a reference
from another problem.

\FloatBarrier

\subsection{Effects of Teacher Supervision on the Student's Predictions}
\label{sec:results:student}

We next measure how the student's next-token predictions change during training. We
denote this change by $\Delta S_c$. In addition to the reference direction
$D_{\mathrm{ref}}$ introduced above, we use the recovery direction
$D_{\mathrm{rec}}$, which compares the predictions of the reference-free teacher with
those of the initial student. For each direction $D$, we compare the cosine similarity
between $\Delta S_c$ and $D$ with the corresponding similarity for the control
direction $D^*$ constructed from another problem.

\begin{figure}[!htb]
\centering
\includegraphics[width=\linewidth]{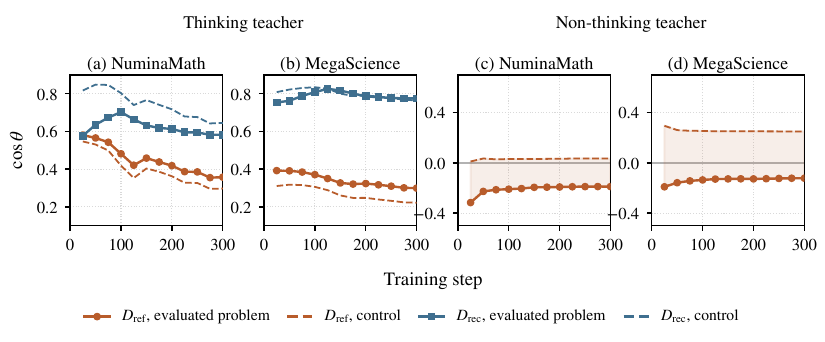}
\caption{Alignment between changes in the Qwen3-8B student's predictions and the
teacher directions during training.}
\label{fig:student-trajectory}
\end{figure}

\FloatBarrier

Under a thinking teacher, Figures~\ref{fig:student-trajectory}(a) and
\ref{fig:student-trajectory}(b) show higher alignment with $D_{\mathrm{rec}}$ than with
$D_{\mathrm{ref}}$. The directions for the evaluated problem and their controls follow
similar trajectories on both datasets. The higher alignment with
$D_{\mathrm{rec}}$ therefore also occurs when the recovery direction is constructed
from another problem. Under a non-thinking teacher,
Figures~\ref{fig:student-trajectory}(c) and
\ref{fig:student-trajectory}(d) show lower alignment with $D_{\mathrm{ref}}$ than with
$D_{\mathrm{ref}}^*$. At step 150, the difference between these cosine similarities is
$-0.23\pm0.03$ on NuminaMath and $-0.38\pm0.04$ on MegaScience.

\begin{figure}[ht]
\centering
\includegraphics[width=\linewidth]{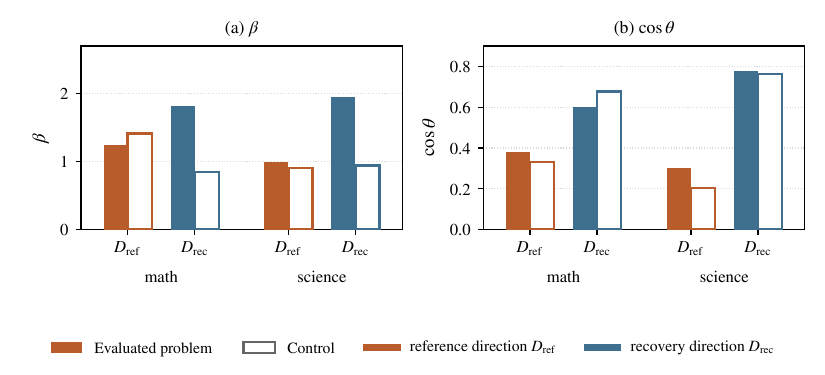}
\caption{Changes in the Qwen3-8B student's predictions measured against the teacher
directions and their controls.
(a) Projection coefficients onto $D_{\mathrm{ref}}$ and $D_{\mathrm{rec}}$ under a
thinking teacher.
(b) Cosine similarity with the same directions.}
\label{fig:student-controls}
\end{figure}

Figure~\ref{fig:student-controls}(a) reports larger projection coefficients for
$D_{\mathrm{rec}}$. Because a projection coefficient also depends on the magnitude of
the teacher direction, a larger coefficient does not necessarily indicate stronger
alignment. Figure~\ref{fig:student-controls}(b) removes this dependence through cosine
similarity. At 8B, the cosine similarity with $D_{\mathrm{ref}}$ exceeds that with
$D_{\mathrm{ref}}^*$ by 0.05 on NuminaMath and 0.10 on MegaScience. The corresponding
differences for $D_{\mathrm{rec}}$ are $-0.08$ and 0.01. The student therefore shows
only weak additional alignment with $D_{\mathrm{ref}}$ beyond the alignment reproduced
by its control, and no consistent additional alignment with $D_{\mathrm{rec}}$.

Table~\ref{tab:rsq} reports $R_c^2$, the fraction of
$\lVert\Delta S_c\rVert^2$ that lies in the span of the teacher directions. For
directions constructed from the evaluated problem, $R_c^2$ ranges from 0.88 to 0.97.
For the control directions, it ranges from 0.94 to 0.98. The control values match or
exceed the values for the evaluated problem under both teacher modes and on both
datasets. A high value of $R_c^2$ therefore does not indicate dependence on the correct
problem context.

\begin{table}[ht]
\centering
\caption{Fraction $R_c^2$ of the squared change in the Qwen3-8B student's log
probabilities that lies in the span of the teacher directions at checkpoint 300.}
\label{tab:rsq}
\small
\begin{tabular*}{0.75\linewidth}{@{\extracolsep{\fill}}llrr@{}}
\toprule
\textbf{Teacher} & \textbf{Dataset} & \textbf{Evaluated problem} & \textbf{Control} \\
\midrule
Thinking     & NuminaMath  & 0.94 & 0.96 \\
Thinking     & MegaScience & 0.97 & 0.98 \\
Non-thinking & NuminaMath  & 0.88 & 0.94 \\
Non-thinking & MegaScience & 0.94 & 0.96 \\
\bottomrule
\end{tabular*}
\end{table}

The corresponding results at 4B and 1.7B appear in
Table~\ref{tab:appendix-scale-diagnostics}. Under a thinking teacher, across the three
tested model sizes, the difference
$\cos\theta_{D_{\mathrm{ref}}}^{(c)}
-\cos\theta_{D_{\mathrm{ref}}^*}^{(c)}$ increases in the order 8B, 4B, and 1.7B on both
datasets. This ordering describes the three models tested here, but does not establish
that model size caused the differences. On NuminaMath, the larger cosine differences at
4B and 1.7B coincide with
higher performance for the student trained with the reference. On MegaScience, the
cosine difference is largest at 1.7B, but the student trained without the reference
performs better on all three science benchmarks. A larger cosine difference therefore
does not reliably indicate a larger performance benefit from the reference.

Under a non-thinking teacher, the cosine similarity between $\Delta S_c$ and
$D_{\mathrm{ref}}$, before comparison with its control, changes sign across model
scales. It is positive at 4B on both datasets and negative or near zero at 8B and
1.7B. However, at 8B, 4B, and 1.7B on both NuminaMath and MegaScience,
$D_{\mathrm{ref}}$ aligns less with $\Delta S_c$ than $D_{\mathrm{ref}}^*$. The sign
of the cosine similarity with $D_{\mathrm{ref}}$ therefore does not determine whether
$D_{\mathrm{ref}}$ aligns more strongly than its control.

\FloatBarrier

\section{Limitations}
\label{sec:limitations}

Our study has several limitations. Our experiments use Qwen3 models at 8B, 4B, and
1.7B. Whether the results generalize to other model families remains unknown. Our
training datasets cover mathematical and scientific reasoning, so the results may not
generalize to coding or other domains. We train each condition once, so our results do
not measure variation across repeated training runs.

\section{Conclusion}
\label{sec:conclusion}

We investigated what privileged reference information contributes to OPSD. We developed
an analysis framework that separates the supervision induced by the reference from
the remaining teacher supervision and tests both against controls constructed from other
problems. Across Qwen3-8B, Qwen3-4B, and Qwen3-1.7B students trained on NuminaMath and
MegaScience, we evaluate the performance benefit of the reference, its alignment with
changes in the student's predictions, and whether either depends on using the reference
for the correct problem.

The experiments provide three forms of evidence. First, students can improve without the
correct reference, and a solution from another problem can outperform the correct
solution. Whether the correct reference improves performance depends on the teacher's
generation mode, the model size, and the training dataset. Second, we compare the change
in the student's predictions during training with the supervision induced by the
reference and with the base model's thinking behavior. The student's predictions align
more strongly with the base model's thinking behavior than with the supervision induced
by the reference. Because controls constructed from other problems reproduce much of the
alignment with both forms of supervision, the stronger alignment with thinking behavior
does not establish that recovering this behavior caused the performance gains. Third,
the alignment gained by using the correct reference instead of a reference from another
problem does not imply a greater performance benefit. Among the tested model sizes and
training datasets, this increase in alignment is largest at 1.7B on MegaScience.
Nevertheless, the student trained without the reference performs better. Together, the
performance experiments and distributional analysis show that the contribution of
privileged reference information to OPSD cannot be inferred from performance gains or
distributional alignment alone.

\section*{Acknowledgements}

We thank Nate Woodward, Collin Farquhar, Shaghayegh Sadeghi, and Nawar Ismail for
their feedback and suggestions on earlier drafts of this paper.

\bibliography{refs}
\bibliographystyle{iclr2025_conference}

\clearpage
\appendix
\section{Additional Experimental Details}
\label{app:details}

\subsection{Dataset Construction}
\label{app:datasets}

\paragraph{NuminaMath.}
We construct the NuminaMath training set from the training split of
NuminaMath-CoT~\citep{numinamath}. We retain problems with an extractable boxed answer
and sample 10,000 examples using the source quotas in
Table~\ref{tab:appendix-data-composition}. Sampling within each source and the final
shuffle use seed 42. We remove exact overlaps with AIME 2024 and MATH-500 after
lowercasing the problem text and collapsing whitespace. AIME 2025 and HMMT 2025
postdate the NuminaMath snapshot and are not included in the decontamination set.

\paragraph{MegaScience.}
We construct the MegaScience training set from the \texttt{textbook\_reasoning} portion
of MegaScience~\citep{fan2025megascience}. We retain solutions that contain an
extractable boxed answer and at least 250 characters. We then sample 10,000 examples
using the subject quotas in Table~\ref{tab:appendix-data-composition}. Sampling uses
seed 42. We remove exact problem matches and shared word 8-grams with GPQA-Diamond,
MMLU-Pro, UGPhysics, SciBench, MATH-500, and GSM8K.

% Non-floating (package-free \captionof equivalent): keeps the table below the section/subsection
% headings and its introductory prose, so it cannot float above the appendix heading.
\begin{center}
\makeatletter\def\@captype{table}\makeatother
\caption{Composition of the training subsets.}
\label{tab:appendix-data-composition}
\small
\setlength{\tabcolsep}{7pt}
\begin{tabular}{llr}
\toprule
\textbf{Dataset} & \textbf{Source or subject} & \textbf{Problems} \\
\midrule
\multirow{4}{*}{NuminaMath}
  & Olympiads & 6,500 \\
  & AoPS forum & 2,000 \\
  & Synthetic AMC & 1,000 \\
  & MATH & 500 \\
\midrule
\multirow{4}{*}{MegaScience}
  & Physics & 4,000 \\
  & Mathematics & 2,500 \\
  & Chemistry & 2,500 \\
  & Biology & 1,000 \\
\bottomrule
\end{tabular}
\end{center}
\FloatBarrier

\paragraph{Comparison of reference types.}
Qwen3-8B and Qwen3-32B generate reasoning traces in thinking mode while receiving the
original dataset solution as an answer anchor. We retain traces whose final boxed
answer agrees with the original answer and whose rendered teacher prompt contains at
most 15,900 tokens. The common set is the deduplicated intersection of the surviving
8B and 32B trace sets. The agreement and length filters favor problems for which both
models produce a correct trace within the limit. Results on this set are therefore not
directly comparable to results on the full 10,000-problem set.

Table~\ref{tab:appendix-matched-data} gives the final artifact counts. The original
solution, two reasoning traces, and answer-only condition use the same 4,953 problems
in the same order. The hint condition contains 4,898 of these problems because 55
answer-leaking hints are removed.

\begin{table}[!ht]
\centering
\caption{Training-set sizes for the reference-type comparison.}
\label{tab:appendix-matched-data}
\small
\setlength{\tabcolsep}{7pt}
\begin{tabular}{lr}
\toprule
\textbf{Reference condition} & \textbf{Problems} \\
\midrule
Original dataset solution & 4,953 \\
Qwen3-8B reasoning trace & 4,953 \\
Qwen3-32B reasoning trace & 4,953 \\
Answer-only reference & 4,953 \\
Abstract hint & 4,898 \\
\bottomrule
\end{tabular}
\end{table}
\FloatBarrier

\subsection{Reference Conditions and Prompts}
\label{app:references}

Table~\ref{tab:appendix-reference-conditions} summarizes the information provided to
the teacher. In every condition except the no-reference condition, the selected text
occupies the reference field in the same teacher prompt. The student does not receive
this text.

\begin{table}[!ht]
\centering
\caption{Reference conditions used during training.}
\label{tab:appendix-reference-conditions}
\small
\setlength{\tabcolsep}{5pt}
\begin{tabularx}{\linewidth}{@{}lX@{}}
\toprule
\textbf{Condition} & \textbf{Information provided to the teacher} \\
\midrule
Original dataset solution
  & The complete solution supplied by the training dataset. \\
Answer-only
  & Only the final answer extracted from the last
    \(\backslash\)\texttt{boxed\{\}} expression in the original solution. \\
Abstract hint
  & Method-level guidance generated from the problem and original solution without a
    worked derivation or final answer. \\
Qwen3-8B trace
  & A verified full reasoning trace generated by Qwen3-8B in thinking mode. \\
Qwen3-32B trace
  & A verified full reasoning trace generated by Qwen3-32B in thinking mode. \\
Mismatched reference
  & The original dataset solution from another problem. \\
No reference
  & No additional text; the teacher receives only the problem. \\
\bottomrule
\end{tabularx}
\end{table}
\FloatBarrier

\paragraph{Generated references.}
The abstract hints are generated by Qwen3-32B in non-thinking mode at temperature 0.7
with top-\(p\) 0.8, top-\(k\) 20, and a 512-token limit. The model is prompted to
produce three to five concise bullets. Hints that contain
\(\backslash\)\texttt{boxed} or reproduce the extracted answer are removed. The
Qwen3-8B and Qwen3-32B traces are generated in thinking mode at temperature 0.6 with
top-\(p\) 0.95, top-\(k\) 20, and a 32,768-token limit. All generated-reference
procedures use seed 42.

\paragraph{Abstract-hint generation prompt.}
Before applying the Qwen3-32B chat template, we construct the following user message:
\begin{quote}
\small\ttfamily
You are a math tutor. Given a problem and a reference solution, write an abstract hint
that helps a student solve the problem without revealing the solution.

\medskip
The hint should mention the core idea, theorem, transformation, invariant, case split,
or caution needed to solve the problem. It should be specific and useful, but not a
full solution.

\medskip
Rules:

- Do not reveal the final answer.

- Do not use \textbackslash boxed\{\}.

- Do not reproduce the solution steps.

- Avoid long calculations or decisive intermediate numeric values.

- Keep it to 3-5 concise bullets.

\medskip
Problem:

\{problem\}

\medskip
Reference solution:

\{solution\}

\medskip
Abstract hint:
\end{quote}
We replace the placeholders with the problem and original dataset solution, render the
prompt as a single user message, and disable thinking mode.

\paragraph{Mismatched references.}
For NuminaMath, we apply a seed-42 derangement to the solution column with no
self-pairs. For MegaScience, we apply the same construction within each subject. The
problem and all other fields remain unchanged.

\paragraph{OPSD training prompts.}
The student user message is
\begin{quote}
\small\ttfamily
Problem: \{problem\}

\medskip
Please reason step by step, and put your final answer within
\textbackslash boxed\{\}.
\end{quote}
When a reference is used, the teacher user message is
\begin{quote}
\small\ttfamily
Problem: \{problem\}

\medskip
Here is a reference solution to this problem:

=== Reference Solution Begin ===

\{reference\}

=== Reference Solution End ===

\medskip
After reading the reference solution above, make sure you truly understand the
reasoning behind each step \textemdash{} do not copy or paraphrase it. Now, using your own words
and independent reasoning, derive the same final answer to the problem above. Think
step by step, explore different approaches, and don't be afraid to backtrack or
reconsider if something doesn't work out:

\medskip
Please reason step by step, and put your final answer within
\textbackslash boxed\{\}.
\end{quote}
In the no-reference condition, the teacher receives the student user message. We
render both messages with the Qwen3 chat template and set thinking mode independently
for the student and teacher.

\subsection{Training Configuration}
\label{app:training-config}

The teacher is the frozen Qwen3-8B base model. During teacher scoring, the student's
LoRA adapter is disabled. We minimize the full-vocabulary forward KL divergence
\(\operatorname{KL}(\pi_T\Vert\pi_S)\). Both distributions are computed at temperature
1.1. Pointwise clipping is applied to each vocabulary coordinate's contribution before
summation, masking, and reduction. We list the common hyperparameters in
Table~\ref{tab:appendix-training-config} and the dataset- and mode-specific settings in
Table~\ref{tab:appendix-regime-config}.

\begin{table}[!ht]
\centering
\caption{Shared OPSD training configuration.}
\label{tab:appendix-training-config}
\small
\setlength{\tabcolsep}{6pt}
\begin{tabular}{lr}
\toprule
\textbf{Setting} & \textbf{Value} \\
\midrule
Base model & Qwen3-8B \\
Student mode & Non-thinking \\
LoRA rank / alpha & 64 / 128 \\
LoRA target modules
  & \(\mathtt{q,k,v,o,gate,up,down\_proj}\) \\
Learning rate & \(5\times10^{-6}\) \\
Optimizer & Fused AdamW \\
Adam betas / epsilon & \(0.9, 0.999\) / \(10^{-8}\) \\
Weight decay / warmup & 0 / 0 \\
Learning-rate schedule & Linear \\
Maximum gradient norm & 0.1 \\
Effective batch size & 32 \\
Training steps & 300 \\
Checkpoint interval & 25 steps \\
Training and data-shuffle seed & 42 \\
Student rollout temperature & 1.1 \\
Student rollout top-\(p\) / top-\(k\) & 0.95 / 20 \\
Student rollout context limit & 20,000 \\
Precision & bfloat16 \\
Distributed training & DeepSpeed ZeRO-2 with CPU optimizer offload \\
\bottomrule
\end{tabular}
\end{table}
\FloatBarrier

The standard microbatch contains two examples per GPU with two gradient-accumulation
steps across eight GPUs. The full-trace conditions use one example per GPU with four
accumulation steps. We use FlashAttention 2, gradient checkpointing, and colocated vLLM
generation with tensor parallel size one. The vLLM rollout engines use their
data-parallel process indices as seeds.

The 20,000-token setting limits the student rollout context. Student prompts are
truncated to leave room for the maximum completion. Teacher prompts are separately
truncated at 20,000 tokens before the sampled student completion is appended. It is
therefore not a universal limit on the combined teacher sequence. For the full-trace
conditions, the 15,900-token prompt filter and 4,096-token completion limit keep the
combined sequence below 20,000 tokens.

\begin{table}[!ht]
\centering
\caption{Settings that vary by dataset and teacher mode.}
\label{tab:appendix-regime-config}
\small
\setlength{\tabcolsep}{6pt}
\begin{tabular}{llrr}
\toprule
\textbf{Dataset} & \textbf{Teacher mode} & \textbf{Completion limit} & \textbf{KL clip per coordinate} \\
\midrule
NuminaMath & Non-thinking & 4,096 & \(10^{-7}\) \\
NuminaMath & Thinking & 1,024 & 0.06 \\
MegaScience & Non-thinking & 2,048 & \(10^{-7}\) \\
MegaScience & Thinking & 1,024 & 0.06 \\
\bottomrule
\end{tabular}
\end{table}
\FloatBarrier

\subsection{Evaluation}
\label{app:evaluation}

We define Avg@\(k\) as the mean accuracy across \(k\) sampled responses. Unparseable
responses count as incorrect. Table~\ref{tab:appendix-evaluation} gives the main
reported read and grader for each benchmark.

\begin{table}[!ht]
\centering
\caption{Evaluation benchmarks and grading procedures.}
\label{tab:appendix-evaluation}
\small
\setlength{\tabcolsep}{4pt}
\begin{tabularx}{\linewidth}{@{}lrlX@{}}
\toprule
\textbf{Benchmark} & \textbf{Problems} & \textbf{Reported read} & \textbf{Grading} \\
\midrule
MATH-500 & 500 & Avg@4
  & Final boxed answer with symbolic equivalence. \\
AIME 2024 & 30 & Avg@12
  & Final boxed answer with symbolic equivalence. \\
AIME 2025 & 30 & Avg@12
  & Final boxed answer with symbolic equivalence. \\
HMMT 2025 & 30 & Avg@12
  & Final boxed answer with symbolic equivalence. \\
GPQA-Diamond & 198 & Avg@12
  & Parsed answer choice with exact letter matching. \\
MMLU-Pro & 12,032 & Avg@1
  & Parsed answer choice with exact letter matching. \\
UGPhysics & 5,520 & Avg@1
  & Official type-aware rule grader at precision 0.01. \\
\bottomrule
\end{tabularx}
\end{table}
\FloatBarrier

The reference-type comparison uses Avg@12 on AIME 2024. Every trained student is
evaluated in non-thinking mode at temperature 1.0 and top-\(p\) 0.8. Students trained
with a non-thinking teacher use repetition penalty 1.3, while students trained with a
thinking teacher use 1.0. Base non-thinking evaluations use repetition penalty 1.0.
Separate base-model evaluations in thinking mode use temperature 1.0 and top-\(p\)
0.95.

\paragraph{Uncertainty estimation.}
We compare two students by resampling the same benchmark problems for both students
10,000 times and recomputing the difference in Avg@\(k\). The paired 95\% interval
contains the 2.5th through 97.5th percentiles of these differences. In
Table~\ref{tab:appendix-reference-type-performance}, each reference type is compared
with the canonical solution. In
Table~\ref{tab:appendix-other-problem-performance}, the solution from another problem
is compared with the correct solution. No other pairs are tested.

\begin{table}[!ht]
\centering
\caption{Reference-type comparison on AIME 2024 under a non-thinking teacher
(Avg@12).}
\label{tab:appendix-reference-type-performance}
\small
\setlength{\tabcolsep}{4pt}
\begin{tabularx}{\linewidth}{@{}Xrrr@{}}
\toprule
\textbf{Reference provided to the teacher} & \textbf{Avg@12} & \textbf{Difference from canonical} & \textbf{Paired 95\% interval} \\
\midrule
Canonical solution & 61.1 & -- & -- \\
Answer only & 66.7 & \(+5.6\) & \([-1.4,+12.8]\) \\
Abstract hint & 31.1 & \(-30.0\) & \([-39.4,-20.6]\) \\
Qwen3-8B reasoning trace & 69.4 & \(+8.3\) & \([+3.3,+13.9]\) \\
Qwen3-32B reasoning trace & 63.3 & \(+2.2\) & \([-2.5,+6.7]\) \\
\bottomrule
\end{tabularx}
\end{table}

\begin{table}[!ht]
\centering
\caption{Correct and mismatched references under a non-thinking teacher (Avg@12).}
\label{tab:appendix-other-problem-performance}
\small
\setlength{\tabcolsep}{4pt}
\begin{tabularx}{\linewidth}{@{}Xrrrr@{}}
\toprule
\textbf{Benchmark} & \textbf{Correct solution} & \textbf{Solution from another problem} & \textbf{Difference} & \textbf{Paired 95\% interval} \\
\midrule
AIME 2024 & 66.7 & 71.9 & \(+5.3\) & \([-0.6,+11.1]\) \\
AIME 2025 & 49.2 & 61.7 & \(+12.5\) & \([+6.4,+18.9]\) \\
HMMT 2025 & 30.6 & 42.2 & \(+11.7\) & \([+6.1,+17.5]\) \\
\bottomrule
\end{tabularx}
\end{table}
\FloatBarrier

\paragraph{Checkpoint selection.}
For each trained condition, we evaluate saved checkpoints on two selection benchmarks
using Avg@4 and select the checkpoint with the highest mean across them. We use
MATH-500 and AIME 2024 for NuminaMath, and MATH-500 and GPQA-Diamond for MegaScience.
The remaining benchmarks are evaluated only after checkpoint selection.
Table~\ref{tab:appendix-checkpoints} gives the selected Qwen3-8B checkpoints.

\begin{table}[!ht]
\centering
\caption{Checkpoints used for the main benchmark comparison.}
\label{tab:appendix-checkpoints}
\small
\setlength{\tabcolsep}{6pt}
\begin{tabular}{llr}
\toprule
\textbf{Training data} & \textbf{Condition} & \textbf{Checkpoint} \\
\midrule
NuminaMath & Non-thinking teacher, correct reference & 50 \\
NuminaMath & Non-thinking teacher, mismatched reference & 50 \\
NuminaMath & Thinking teacher, correct reference & 200 \\
NuminaMath & Thinking teacher, no reference & 200 \\
\midrule
MegaScience & Non-thinking teacher, correct reference & 150 \\
MegaScience & Non-thinking teacher, mismatched reference & 150 \\
MegaScience & Thinking teacher, correct reference & 300 \\
MegaScience & Thinking teacher, no reference & 250 \\
\bottomrule
\end{tabular}
\end{table}
\FloatBarrier

\paragraph{Generation limits.}
Table~\ref{tab:appendix-eval-limits} reports the maximum generated tokens in the
main performance evaluations. Every condition compared within a benchmark uses the same
completion limit, so a single limit per benchmark applies to all conditions.

\begin{table}[!ht]
\centering
\caption{Generation limits for the main reported evaluations.}
\label{tab:appendix-eval-limits}
\small
\begin{tabularx}{\linewidth}{@{}Xr@{}}
\toprule
\textbf{Benchmark} & \textbf{Completion limit (tokens)} \\
\midrule
MATH-500 & 38,912 \\
AIME 2024, AIME 2025, HMMT 2025 & 32,768 \\
GPQA-Diamond & 32,768 \\
MMLU-Pro & 32,768 \\
UGPhysics & 32,768 \\
\bottomrule
\end{tabularx}
\end{table}
\FloatBarrier

\paragraph{UGPhysics coverage.}
The base, non-thinking-teacher, and mismatched-reference rows use all 5,520 UGPhysics
questions. The thinking-teacher row uses 5,477 questions after a transient
dataset-configuration load failure. The corresponding no-reference row uses 5,175
questions because one 345-problem evaluation shard did not complete before the job
limit.

\subsection{Evaluation in Thinking Mode}
\label{app:thinking-eval}

We additionally evaluate the Qwen3-8B base model and the students trained on
NuminaMath with a thinking teacher, with and without the reference. Thinking mode is
enabled for every model, as in \citet{zhao2026opsd}. We set the maximum generation
length to 38{,}912 tokens, temperature to 1.0, top-$p$ to 0.95, top-$k$ to $-1$, and
the repetition penalty to 1.0. Table~\ref{tab:thinking-eval} reports Avg@12 accuracy on
AIME 2024, AIME 2025, and HMMT 2025. The checkpoints were selected using the main
non-thinking evaluation.

% GENERATED by scripts/gen_table_thinking_eval.py from data/thinking_eval_8b.json. Do not edit by hand.
\begin{table}[t]
\centering
\caption{Thinking-mode evaluation of Qwen3-8B on mathematical reasoning benchmarks.}
\label{tab:thinking-eval}
\small
\begin{tabular}{lrrrr}
\toprule
\textbf{Method} & \textbf{AIME24} & \textbf{AIME25} & \textbf{HMMT25} & \textbf{Avg.} \\
\midrule
\rowcolor{gray!10}
Base model & 74.72 & 67.78 & 45.00 & 62.50 \\
OPSD, thinking teacher, correct ref. & 75.28 & 70.56 & 47.50 & 64.45 \\
OPSD, thinking teacher, no ref. & 76.67 & 71.39 & 45.00 & 64.35 \\
\bottomrule
\end{tabular}
\end{table}

\subsection{Software and Compute}
\label{app:software}

We train on one node with eight NVIDIA H200 GPUs. Each training condition uses one
training run. Primary launchers request 64 CPU cores and 256 GB of host memory.
Table~\ref{tab:appendix-software} gives the principal software versions from the
training environment.

\begin{table}[!ht]
\centering
\caption{Principal software versions.}
\label{tab:appendix-software}
\small
\setlength{\tabcolsep}{7pt}
\begin{tabular}{lr@{\qquad}lr}
\toprule
\textbf{Package} & \textbf{Version} & \textbf{Package} & \textbf{Version} \\
\midrule
Python & 3.10.12 & PyTorch & 2.8.0 \\
Transformers & 4.57.1 & TRL & 0.26.0 \\
PEFT & 0.17.1 & DeepSpeed & 0.18.2 \\
Datasets & 3.6.0 & Accelerate & 1.11.0 \\
vLLM & 0.11.0 & FlashAttention & 2.8.3 \\
math-verify & 0.8.0 & & \\
\bottomrule
\end{tabular}
\end{table}
\FloatBarrier

\section{Additional Distributional Analyses}
\label{app:distributional}

This section reports additional measurements of the teacher's signal and the student's
change. The change-magnitude and alignment analyses use the fixed responses generated by
the initial student, following Section~\ref{sec:setup:fixed-text}. The policy-entropy
measurements are instead recorded from the on-policy trajectories generated during
training.

\subsection{Sensitivity to Control Pairing}
\label{app:control-pairings}

Each control pairs an evaluated problem with a different problem. To determine whether
the results depend on this pairing, we repeat each control construction under three
different assignments while keeping the student responses and original measurements
fixed. Table~\ref{tab:appendix-control-pairings} reports the range of the
difference between the evaluated-problem and control cosine similarities across the
three assignments.

\begin{table}[!ht]
\centering
\caption{Cosine-similarity differences across three control assignments. Bracketed
entries show the minimum and maximum values, not confidence intervals.}
\label{tab:appendix-control-pairings}
\small
\setlength{\tabcolsep}{4pt}
\begin{tabularx}{\linewidth}{@{}lXcc@{}}
\toprule
\textbf{Analysis} & \textbf{Comparison} & \textbf{NuminaMath} & \textbf{MegaScience} \\
\midrule
Figure~\ref{fig:signal}
  & Canonical solution & $[-0.146,-0.133]$ & -- \\
Figure~\ref{fig:signal}
  & Qwen3-8B trace & $[-0.069,-0.062]$ & -- \\
Figure~\ref{fig:signal}
  & Qwen3-32B trace & $[-0.081,-0.068]$ & -- \\
Figure~\ref{fig:signal}
  & Abstract hint & $[-0.144,-0.136]$ & -- \\
Figure~\ref{fig:signal}
  & Answer only & $[-0.150,-0.145]$ & -- \\
\midrule
Figure~\ref{fig:student-trajectory}
  & Reference direction at step 150 & $[0.054,0.055]$ & $[0.066,0.079]$ \\
Figure~\ref{fig:student-trajectory}
  & Recovery direction at step 150 & $[-0.135,-0.129]$ & $[0.015,0.017]$ \\
\midrule
Figure~\ref{fig:student-controls}(b)
  & Reference direction & $[0.045,0.048]$ & $[0.077,0.099]$ \\
Figure~\ref{fig:student-controls}(b)
  & Recovery direction & $[-0.077,-0.073]$ & $[0.014,0.018]$ \\
\bottomrule
\end{tabularx}
\end{table}

The sign of every cosine difference is unchanged across the three assignments. The
magnitude-sensitive reference projection in Figure~\ref{fig:student-controls}(a) is less
stable on MegaScience, where the difference between the evaluated-problem and control
measurements ranges from $-0.038$ to $0.088$. Because the projection coefficient also
depends on the direction's magnitude, we do not interpret this variation as a stable
directional effect. The
corresponding cosine difference remains positive across all three assignments.

\subsection{Change Magnitude and Recovery Divergence}
\label{app:magnitude}

\begin{figure}[t]
\centering
\includegraphics[width=\linewidth]{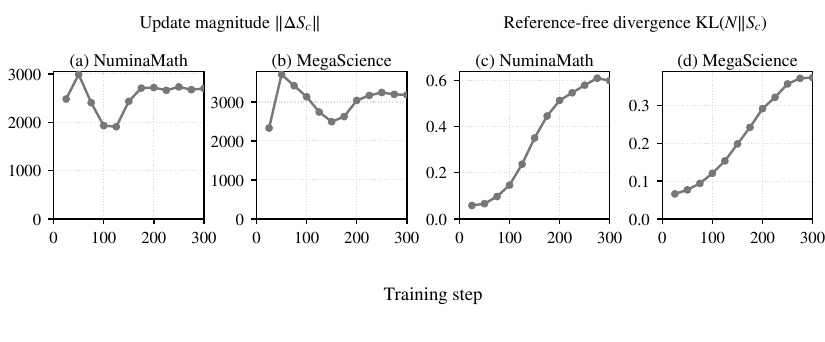}
\caption{Magnitude of the Qwen3-8B student's log-probability change and its KL
divergence from the reference-free teacher during training under a thinking teacher.}
\label{fig:appendix-magnitude}
\end{figure}

As Figure~\ref{fig:appendix-magnitude} shows, after the initial variation,
$\lVert\Delta S_c\rVert_{\mathcal{I}_0}$ remains within a relatively narrow range on
both datasets, while $\operatorname{KL}_{\mathcal{I}_0}(N\Vert S_c)$ continues to
increase.
The student therefore becomes increasingly different from the reference-free teacher
without a corresponding increase in the overall magnitude of its log-probability
change.

\subsection{Policy Entropy}
\label{app:entropy}

\begin{figure}[t]
\centering
\includegraphics[width=0.65\linewidth]{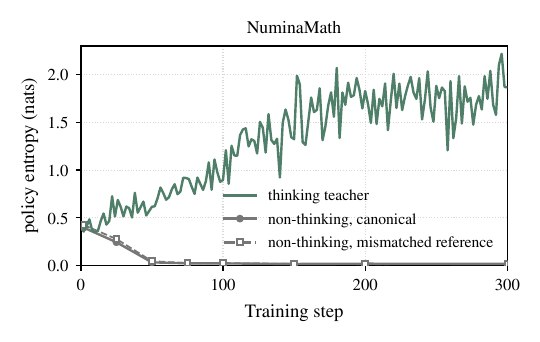}
\caption{Policy entropy of the Qwen3-8B student during training on NuminaMath under
thinking and non-thinking teachers.}
\label{fig:appendix-entropy}
\end{figure}

Figure~\ref{fig:appendix-entropy} shows that entropy increases under a thinking teacher
but falls sharply under a non-thinking teacher. The near-identical non-thinking curves
show that replacing the reference does not explain the entropy collapse.

\subsection{Alignment with the Thinking Direction}
\label{app:thinking-direction}

\begin{figure}[t]
\centering
\includegraphics[width=\linewidth]{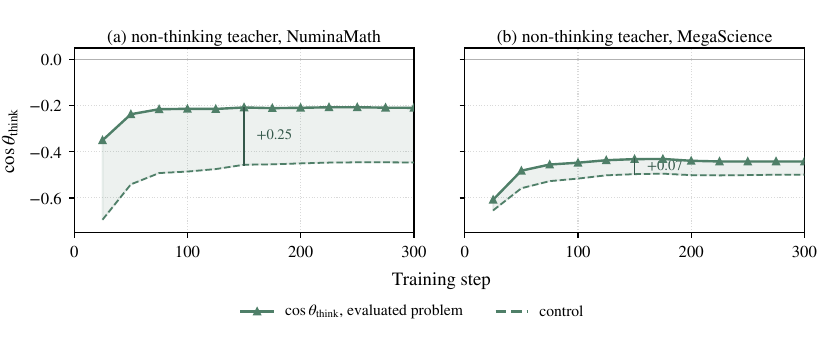}
\caption{Cosine similarity between changes in the Qwen3-8B student's predictions and
the base model's thinking direction under a non-thinking teacher.}
\label{fig:appendix-thinking-direction}
\end{figure}

In Figure~\ref{fig:appendix-thinking-direction}, both the evaluated-problem and control
measurements remain negatively aligned with the thinking direction throughout training.
The evaluated-problem measurement is less negative than its control, especially on
NuminaMath.
Thus, the positive difference between the two measurements reflects weaker opposition
to the thinking direction rather than positive alignment with it.

\FloatBarrier

\section{Results at Smaller Model Scales}
\label{app:scale}

\subsection{Qwen3-4B Performance}
\label{app:scale-4b}

Table~\ref{tab:appendix-4b-performance} repeats the main performance comparison with
Qwen3-4B using the same training and evaluation procedure as for Qwen3-8B. On
NuminaMath, the correct reference improves the thinking-teacher student on all three
competition mathematics benchmarks, with paired
bootstrap intervals excluding zero. Under a non-thinking teacher, the mismatched
reference outperforms the correct reference on the same benchmarks. On MegaScience,
removing the reference does not reduce thinking-teacher performance.

\FloatBarrier
\begin{table}[!ht]
\centering
\caption{Qwen3-4B benchmark results under non-thinking evaluation. Evaluation reads
follow Table~\ref{tab:main}.}
\label{tab:appendix-4b-performance}
\vspace{7pt}
\footnotesize
\setlength{\tabcolsep}{1.5pt}
\renewcommand{\arraystretch}{1.04}
\begin{tabularx}{\linewidth}{@{}l
  >{\raggedleft\arraybackslash\hsize=1.174\hsize}X
  >{\raggedleft\arraybackslash\hsize=0.982\hsize}X
  >{\raggedleft\arraybackslash\hsize=1.220\hsize}X
  >{\raggedleft\arraybackslash\hsize=1.146\hsize}X
  >{\raggedleft\arraybackslash\hsize=0.478\hsize}X@{}}
\toprule
\textbf{Method} &
\multicolumn{5}{c}{\textbf{NuminaMath}} \\
\cmidrule(lr){2-6}
 &
\textbf{MATH-500} & \textbf{AIME24} & \textbf{AIME25} &
\textbf{HMMT25} & \textbf{Avg.} \\
\midrule
\rowcolor{gray!10}
Base model & 84.0 & 24.7 & 21.1 & 10.8 & 35.2 \\
\addlinespace[2pt]
OPSD, non-thinking teacher, correct ref. & 82.5 & 31.4 & 21.7 & 13.6 & 37.3 \\
OPSD, non-thinking teacher, mismatched ref. & 82.6 & 40.3 & 32.5 & 22.8 & 44.5 \\
OPSD, thinking teacher, correct ref. & 90.9 & 56.7 & 48.3 & 34.2 & 57.5 \\
OPSD, thinking teacher, no ref. & 90.0 & 41.9 & 37.5 & 24.4 & 48.5 \\
\midrule
\textbf{Method} &
\multicolumn{5}{c}{\textbf{MegaScience}} \\
\cmidrule(lr){2-6}
 &
\textbf{MATH-500} & \textbf{GPQA-D} & \textbf{MMLU-Pro} &
\textbf{UGPhysics} & \textbf{Avg.} \\
\midrule
\rowcolor{gray!10}
Base model & 84.0 & 43.5 & 62.0 & 12.8 & 50.6 \\
\addlinespace[2pt]
OPSD, non-thinking teacher, correct ref. & 81.1 & 42.6 & 58.1 & 12.7 & 48.6 \\
OPSD, non-thinking teacher, mismatched ref. & 81.0 & 44.4 & 57.8 & 12.8 & 49.0 \\
OPSD, thinking teacher, correct ref. & 93.6 & 49.2 & 64.7 & 26.3 & 58.4 \\
OPSD, thinking teacher, no ref. & 92.8 & 51.6 & 67.7 & 26.0 & 59.5 \\
\bottomrule
\end{tabularx}
\end{table}
\FloatBarrier

\subsection{Qwen3-1.7B Performance}
\label{app:scale-1p7b}

Table~\ref{tab:appendix-1p7b-performance} repeats the comparison with Qwen3-1.7B under
the same evaluation protocol. On NuminaMath, the correct reference improves the
thinking-teacher student on AIME 2024 and HMMT 2025, with paired bootstrap intervals
excluding zero; the interval for
AIME 2025 includes zero. Under a non-thinking teacher, the mismatched reference
outperforms the correct reference on all three competition mathematics benchmarks.
On MegaScience, removing the reference improves thinking-teacher performance on all
three science benchmarks.

\FloatBarrier
\begin{table}[!ht]
\centering
\caption{Qwen3-1.7B benchmark results under non-thinking evaluation. Evaluation reads
follow Table~\ref{tab:main}.}
\label{tab:appendix-1p7b-performance}
\vspace{7pt}
\footnotesize
\setlength{\tabcolsep}{1.5pt}
\renewcommand{\arraystretch}{1.04}
\begin{tabularx}{\linewidth}{@{}l
  >{\raggedleft\arraybackslash\hsize=1.174\hsize}X
  >{\raggedleft\arraybackslash\hsize=0.982\hsize}X
  >{\raggedleft\arraybackslash\hsize=1.220\hsize}X
  >{\raggedleft\arraybackslash\hsize=1.146\hsize}X
  >{\raggedleft\arraybackslash\hsize=0.478\hsize}X@{}}
\toprule
\textbf{Method} &
\multicolumn{5}{c}{\textbf{NuminaMath}} \\
\cmidrule(lr){2-6}
 &
\textbf{MATH-500} & \textbf{AIME24} & \textbf{AIME25} &
\textbf{HMMT25} & \textbf{Avg.} \\
\midrule
\rowcolor{gray!10}
Base model & 72.0 & 14.4 & 10.3 & 4.4 & 25.3 \\
\addlinespace[2pt]
OPSD, non-thinking teacher, correct ref. & 67.2 & 13.1 & 11.9 & 5.3 & 24.4 \\
OPSD, non-thinking teacher, mismatched ref. & 77.5 & 27.5 & 18.3 & 13.9 & 34.3 \\
OPSD, thinking teacher, correct ref. & 76.5 & 32.2 & 20.8 & 14.7 & 36.1 \\
OPSD, thinking teacher, no ref. & 79.1 & 22.2 & 16.7 & 10.0 & 32.0 \\
\midrule
\textbf{Method} &
\multicolumn{5}{c}{\textbf{MegaScience}} \\
\cmidrule(lr){2-6}
 &
\textbf{MATH-500} & \textbf{GPQA-D} & \textbf{MMLU-Pro} &
\textbf{UGPhysics} & \textbf{Avg.} \\
\midrule
\rowcolor{gray!10}
Base model & 72.0 & 30.4 & 46.3 & 8.9 & 39.4 \\
\addlinespace[2pt]
OPSD, non-thinking teacher, correct ref. & 71.4 & 29.0 & 39.4 & 8.4 & 37.0 \\
OPSD, non-thinking teacher, mismatched ref. & 70.0 & 26.8 & 34.4 & 7.9 & 34.8 \\
OPSD, thinking teacher, correct ref. & 81.5 & 33.4 & 50.9 & 13.5 & 44.8 \\
OPSD, thinking teacher, no ref. & 87.6 & 36.7 & 55.7 & 17.2 & 49.3 \\
\bottomrule
\end{tabularx}
\end{table}
\FloatBarrier

\subsection{Distributional Results Across Model Scales}
\label{app:scale-diagnostics}

We repeat the cosine-similarity analysis at 8B, 4B, and 1.7B.
Table~\ref{tab:appendix-scale-diagnostics} reports each cosine similarity and its
difference from the corresponding control across three control assignments. The
checkpoint used for each measurement is listed in the table.

\begin{table}[!ht]
\centering
\caption{Cosine similarities at 8B, 4B, and 1.7B. Bracketed entries give the range
across three control assignments, not confidence intervals.}
\label{tab:appendix-scale-diagnostics}
\vspace{7pt}
\scriptsize
\setlength{\tabcolsep}{3.5pt}
\renewcommand{\arraystretch}{1.04}
\textbf{(a) Thinking teacher}\\[3pt]
\begin{tabular}{@{}lccrrrr@{}}
\toprule
\textbf{Dataset} & \textbf{Scale} & \textbf{Checkpoint} &
\multicolumn{2}{c}{\textbf{Reference direction $D_{\mathrm{ref}}$}} &
\multicolumn{2}{c}{\textbf{Recovery direction $D_{\mathrm{rec}}$}} \\
\cmidrule(lr){4-5}\cmidrule(lr){6-7}
 & & & $\cos\theta_{D_{\mathrm{ref}}}^{(c)}$ &
$\cos\theta_{D_{\mathrm{ref}}}^{(c)}-\cos\theta_{D_{\mathrm{ref}}^*}^{(c)}$ &
$\cos\theta_{D_{\mathrm{rec}}}^{(c)}$ &
$\cos\theta_{D_{\mathrm{rec}}}^{(c)}-\cos\theta_{D_{\mathrm{rec}}^*}^{(c)}$ \\
\midrule
NuminaMath  & 8B & 250 & 0.3785 & $[+0.0447,+0.0479]$ & 0.5989 & $[-0.0773,-0.0727]$ \\
NuminaMath  & 4B & 150 & 0.4020 & $[+0.2161,+0.2308]$ & 0.7910 & $[+0.0206,+0.0219]$ \\
NuminaMath  & 1.7B & 150 & 0.5668 & $[+0.2699,+0.2963]$ & 0.8396 & $[-0.0152,-0.0139]$ \\
MegaScience & 8B & 300 & 0.3037 & $[+0.0775,+0.0994]$ & 0.7787 & $[+0.0140,+0.0176]$ \\
MegaScience & 4B & 300 & 0.4809 & $[+0.1367,+0.1394]$ & 0.8991 & $[+0.0555,+0.0576]$ \\
MegaScience & 1.7B & 100 & 0.5210 & $[+0.3998,+0.4123]$ & 0.9183 & $[+0.0076,+0.0092]$ \\
\bottomrule
\end{tabular}

\vspace{6pt}
\textbf{(b) Non-thinking teacher}\\[3pt]
\begin{tabular}{@{}lccrr@{}}
\toprule
\textbf{Dataset} & \textbf{Scale} & \textbf{Checkpoint} &
\multicolumn{2}{c}{\textbf{Reference direction $D_{\mathrm{ref}}$}} \\
\cmidrule(lr){4-5}
 & & & $\cos\theta_{D_{\mathrm{ref}}}^{(c)}$ &
$\cos\theta_{D_{\mathrm{ref}}}^{(c)}-\cos\theta_{D_{\mathrm{ref}}^*}^{(c)}$ \\
\midrule
NuminaMath  & 8B & 50  & $-0.2272$ & $[-0.2873,-0.2621]$ \\
NuminaMath  & 4B & 50  & $+0.2997$ & $[-0.2550,-0.2392]$ \\
NuminaMath  & 1.7B & 50  & $-0.1410$ & $[-0.2461,-0.2289]$ \\
MegaScience & 8B & 150 & $-0.1276$ & $[-0.3825,-0.3632]$ \\
MegaScience & 4B & 100 & $+0.3556$ & $[-0.2524,-0.2481]$ \\
MegaScience & 1.7B & 100 & $+0.0240$ & $[-0.3907,-0.3525]$ \\
\bottomrule
\end{tabular}
\end{table}

\FloatBarrier

\end{document}